\documentclass[10pt,twocolumn,letterpaper]{article}
\usepackage[pagenumbers]{cvpr}

\usepackage{amsmath,amsfonts,bm}

\def\eqref#1{equation~\ref{#1}}

\def\1{\bm{1}}

\DeclareMathAlphabet{\mathsfit}{\encodingdefault}{\sfdefault}{m}{sl}
\SetMathAlphabet{\mathsfit}{bold}{\encodingdefault}{\sfdefault}{bx}{n}

\usepackage{colortbl}
\usepackage{tabularx}

\definecolor{cvprblue}{rgb}{0.21,0.49,0.74}
\usepackage[pagebackref,breaklinks,colorlinks,allcolors=cvprblue]{hyperref}
\hypersetup{
  pdfauthor={Chun Gu, Xiaofei Wei, Yiming Chen, Juntong Zhang, Li Zhang},
  pdfsubject={Robotics preprint}
}
\title{X-World: Transferring Video Priors to Cross-Embodiment Robot Control}
\hypersetup{pdftitle={\thetitle}}
\newcommand{\method}{\textit{X-World}}
\newcommand{\mainsectionref}[1]{Sec.~\ref{#1}}

\providecommand{\keywords}[1]{}

\definecolor{rankOne}{HTML}{B57FA6}
\definecolor{rankTwo}{HTML}{8A8FC3}
\newcommand{\first}[1]{\cellcolor{rankOne!22}\textbf{#1}}
\newcommand{\second}[1]{\cellcolor{rankTwo!22}\underline{#1}}

\newcommand{\paperfig}[4]{%
  \begin{figure*}[t]
    \centering
    \includegraphics[width=#2]{#3}
    \caption{#4}\label{#1}
  \end{figure*}
}

\author{%
  Chun Gu\textsuperscript{1,2,3},
  Xiaofei Wei\textsuperscript{1,2,3},
  Yiming Chen\textsuperscript{2,3},
  Juntong Zhang\textsuperscript{2,3},
  Li Zhang\textsuperscript{1,2,3$\dagger$}\\[0.4em]
  \normalsize
  \textsuperscript{1}School of Data Science, Fudan University\quad
  \textsuperscript{2}Shanghai Innovation Institute\quad
  \textsuperscript{3}Logos Intelligence Lab\\[0.4em]
  \normalsize\url{https://logosintelligencelab.github.io/X-World/}%
}
\AddToHook{cmd/@maketitle/after}{\noindent\begin{minipage}{\textwidth}
  \centering
  \captionsetup{type=figure}
  \includegraphics[width=\linewidth]{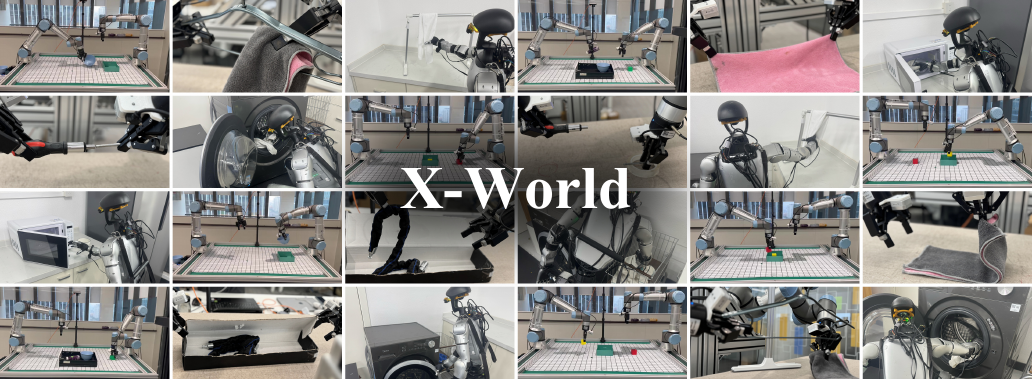}
  \caption{\textbf{Task examples on three real-robot configurations.} Two dual-arm UR5e setups
    illustrate bimanual manipulation, and the Unitree G1 illustrates whole-body manipulation.}
  \label{fig:teaser}
\end{minipage}\par
\vspace{9pt}
}

\begin{document}
\maketitle
\begingroup
  \renewcommand{\thefootnote}{\fnsymbol{footnote}}%
  \footnotetext[2]{Corresponding author: \href{mailto:lizhangfd@fudan.edu.cn}{\nolinkurl{lizhangfd@fudan.edu.cn}}.}
\endgroup

\raggedbottom
\AddToHookNext{shipout/after}{\flushbottom}
\begin{abstract}
Language-instructed manipulation requires robust closed-loop policies, yet action-labeled robot trajectories are costly and limited in coverage.
Human and robot videos provide complementary spatiotemporal priors for cross-embodiment control.
We propose \method{}, which adapts pretrained video diffusion to variable views. A Q-Former compresses hidden states into tokens for a shared flow-matching expert that predicts action chunks.
Dataset-specific projections map native states and actions.
We adapt the backbone on large-scale robot and human videos, freeze it for action-only pretraining of the Q-Former and action expert, then post-train on the target platform.
At inference, \method{} responds at 20\,Hz and overlaps updates with execution via asynchronous real-time chunking.
\method{} performs strongly on LIBERO and under zero-shot shifts on LIBERO-Plus. On hardware, it averages 81.7\% success across six bimanual tasks on the side-mounted dual-arm UR5e, one of three real-robot configurations presented.
\end{abstract}

\section{Introduction}
\label{sec:intro}

Language-instructed robotic manipulation~\cite{brohan2022rt,brohan2023rt} requires policies that translate high-level instructions into robust low-level actions under diverse contacts, partial observability, and long horizons.
Recent vision-language-action (VLA) policies~\cite{brohan2022rt,brohan2023rt,kim2024openvla,team2024octo,bjorck2025gr00t,black2024pi0,pi05} rely on large collections of action-labeled trajectories that are costly to obtain and often limited in task, embodiment, and viewpoint coverage.
Internet-scale videos of humans and robots capture spatiotemporal patterns such as motion and contact.
How can video world priors improve closed-loop manipulation across embodiments with limited action supervision?

Prior work~\cite{ye2024latent,du2023learning,hu2024video} suggests that pretrained video diffusion models can aid action generation. Many methods assume a single view or a fixed camera layout, making them harder to deploy on robots with different numbers of cameras.
Multi-view adaptations~\cite{liao2025genie} can shift the backbone away from its pretrained behavior and weaken transfer.
Across robot embodiments, datasets~\cite{o2024open,bu2025agibot,wu2025robocoin,jiang2025galaxea,wu2024robomind} define states and actions in different coordinate frames, units, and control conventions.
Directly pooling these action targets in one space can produce incompatible supervision and destabilize learning.
Iterative sampling and large backbones can hinder real-time control by increasing latency.

To address these limitations, we propose \method{}, which (i) adapts a pretrained video diffusion backbone (LTX-Video~\cite{hacohen2024ltx}) to variable-view robot observations via view embeddings and a zero-initialized cross-view attention module, (ii) distills diffusion hidden states into a compact token sequence via a Q-Former as a world-prior interface for action generation, (iii) trains a shared action expert across embodiments and, following X-VLA~\cite{xvla}, uses dataset-specific projections to encode states and actions and decode predictions into native actions, and (iv) supports continuous control via asynchronous real-time chunking that decouples inference updates from action execution to avoid waiting latency.
We train \method{} in three steps to separate learning a transferable video world prior from learning control. First, we adapt the pretrained video diffusion backbone on large-scale robot~\cite{o2024open,bu2025agibot,wu2025robocoin,jiang2025galaxea,wu2024robomind} and human videos~\cite{grauman2022ego4d,zhao2025humanoid} to acquire a strong spatiotemporal prior. Then, we freeze the video diffusion backbone and optimize the Q-Former and action expert with the action-only objective $\mathcal{L}_{act}$ for knowledge isolation, so action gradients do not perturb the learned prior. Finally, we post-train on target-platform data to adapt the policy for benchmark or real-world deployment.

We evaluate \method{} on LIBERO~\cite{liu2023libero} and test the same checkpoint under zero-shot distribution shifts on LIBERO-Plus~\cite{fei2025liberoplus}.
The policy performs well on both benchmarks, with its largest gains over OpenVLA-OFT~\cite{OFT} on LIBERO-Plus occurring under robot, language, and layout perturbations.
On real hardware, we evaluate \method{} on six bimanual tasks on the side-mounted dual-arm UR5e.
Figure~\ref{fig:teaser} shows task examples from all three configurations, including textile handling on the shared-frame dual-arm UR5e and interaction with a household appliance on the Unitree G1.

Our contributions are threefold.
\textbf{(i)} We upgrade a pretrained video diffusion backbone to variable-view robot videos via view embeddings and a zero-initialized cross-view attention module.
\textbf{(ii)} We distill diffusion hidden states into a token sequence via a Q-Former and train a shared flow-matching action expert across embodiments, with dataset-specific projections for native state and action spaces.
\textbf{(iii)} We train in three stages by adapting the video diffusion backbone on large-scale videos and learning action generation with the backbone kept frozen for knowledge isolation, and deploy with asynchronous real-time chunking for low-latency control.

\begin{figure*}[t]
\centering
\setlength{\parskip}{0pt}
\setlength{\abovecaptionskip}{4pt}
\includegraphics[width=\linewidth]{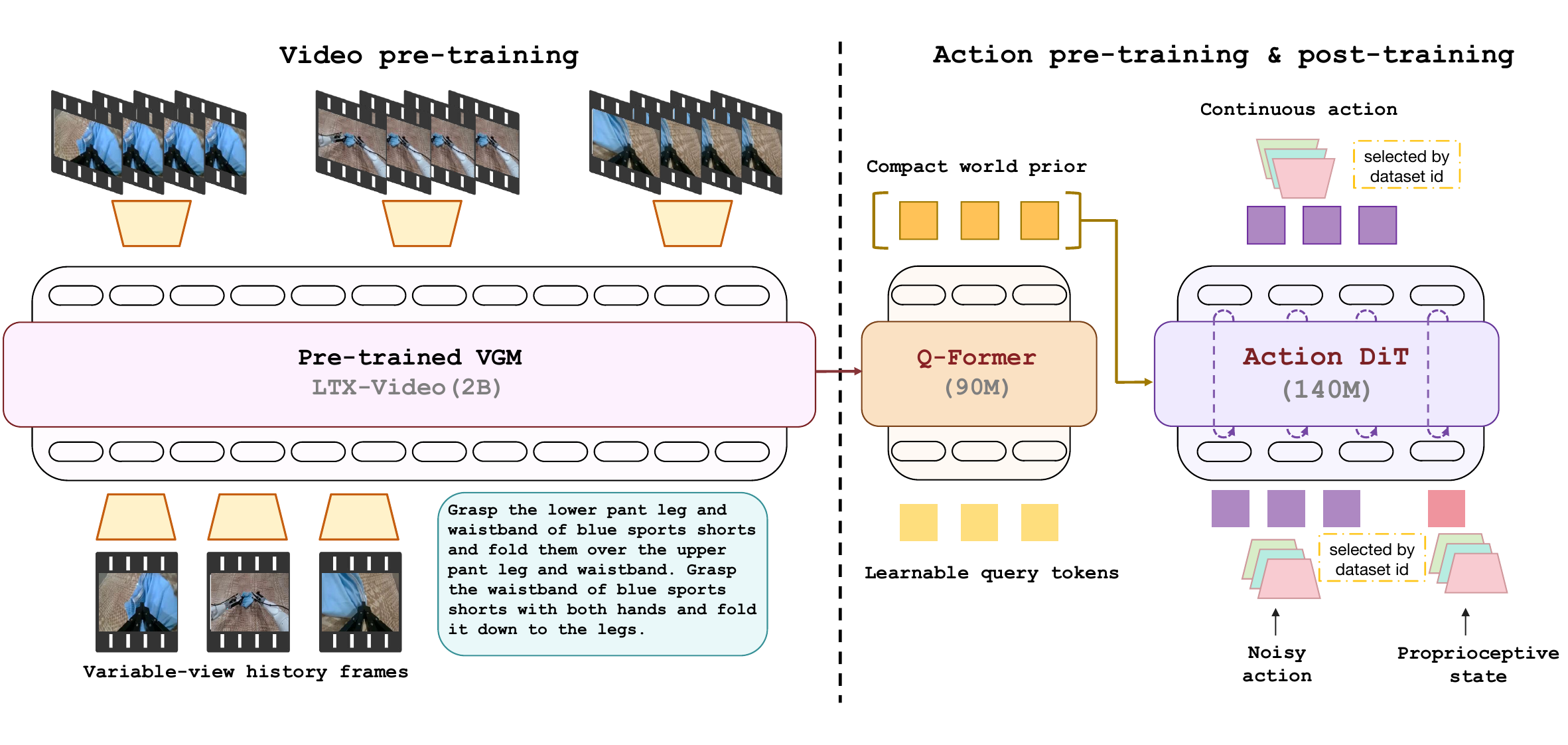}\par\nointerlineskip
\caption{\textbf{Overview of \method{}.} The video backbone is first adapted on robot and human videos with one to three camera views (left). A layer-aligned Q-Former then compresses its features into fixed-length tokens that condition the action DiT, while dataset-specific projections map native states and actions to and from the shared expert (right). Action pretraining learns this interface with the video backbone frozen; target-domain post-training updates all modules using video and action supervision.}\label{fig:pipeline}
\end{figure*}

\section{Related work}
\label{sec:related}

\noindent\textbf{Video-action models.}
Video-action models bridge visual dynamics and control by leveraging world priors inherited from video foundation models for action generation. Recent advancements in this domain~\cite{hu2024video,liao2025genie,li2026causal} have demonstrated capabilities comparable to vision-language-action (VLA) policies across a broad spectrum of tasks. Furthermore, in evaluations targeting generalization and zero-shot task completion, these models have been reported to match or even surpass VLA policies in specific settings~\cite{ye2026world}.
Early frameworks~\cite{du2023learning,du2023video,wen2024vidman} typically adopt a two-stage design, where they first generate future observations as explicit video rollouts and subsequently employ inverse dynamics to map predicted state transitions back to actions. While this pipeline renders the world rollout explicit, prediction errors can propagate to the downstream control module, and the decoupled architecture often increases engineering complexity for closed-loop deployment.
To mitigate these issues, recent works extract internal generative features, such as denoising hidden states or latent representations, to serve as world priors for action generation. Certain methods~\cite{hu2024video,yang2025covar} compress these features into compact tokens for cross-attention with an action expert, whereas others~\cite{bi2025motus,liao2025genie,ye2026world,li2026causal,pai2025mimic} jointly model video and action sequences using mixture-of-transformers architectures.
In parallel, Cosmos Policy~\cite{kim2026cosmos} represents actions as image-like tokens and reuses video generators to synthesize action sequences directly, offering an alternative that avoids a separate action expert.
In contrast, \method{} adapts a pretrained video diffusion backbone to variable-view robot videos and bypasses explicit video rollouts by compressing its per-layer denoising hidden states into a token sequence via a layer-aligned Q-Former, while following a three-phase training pipeline to better preserve the video model priors.

\noindent\textbf{Cross-embodiment training.}
The emergence of robotic foundation models has been significantly propelled by scaling training across expansive datasets from diverse platforms~\cite{o2024open}. However, naive aggregation of trajectories from disparate embodiments often yields diminishing returns, primarily due to inherent discrepancies in action interfaces and observation spaces. A burgeoning line of work~\cite{doshi2024scaling,wang2024scaling,zheng2025universal,xvla} addresses these heterogeneities by introducing alignment layers or adapters that project dataset-specific signals into a canonical representation. Beyond action-state mismatches, observation heterogeneity arising from varied camera extrinsics and robot morphologies poses additional challenges. For instance, VLA-LPAF~\cite{bian2025vla} alleviates viewpoint-induced shifts via representation-level feature alignment, while other strategies operate at the data level through novel-view synthesis~\cite{chen2024rovi} or embodiment inpainting~\cite{chen2024mirage,chen2024rovi,ji2025oxe}. Distinctively, \method{} capitalizes on large-scale video pretraining to internalize world dynamics across diverse observations, while concurrently bridging heterogeneous action interfaces via modular adapters to facilitate unified policy learning.

\section{Method}
\label{sec:method}

\subsection{Overview}
\label{subsec:method-setup}

We study language-instructed closed-loop manipulation where, at time step $t$, the policy takes variable-view RGB observations $\mathcal{I}_t=\{I_t^{(v)}\}_{v=1}^{V_t}$, a language instruction $l$, and proprioceptive state $s_t$, and outputs a short action chunk in the embodiment's native action space.
In Sec.~\ref{subsec:method-arch}, we introduce \method{}, which seamlessly adapts a pretrained video diffusion model to robot observations and leverages its world prior to scale action learning on heterogeneous VLA datasets.
In Sec.~\ref{subsec:method-training}, we describe a three-stage training pipeline that learns the video world prior from videos and transfers it to action generation.
Finally, in Sec.~\ref{subsec:method-inference}, we present asynchronous real-time chunking for low-latency closed-loop control. Figure~\ref{fig:pipeline} shows the full pipeline.

\subsection{Model architecture}
\label{subsec:method-arch}

\method{} consists of three modules: (i) a variable-view video diffusion backbone that encodes spatiotemporal world representations, (ii) a Q-Former that distills them into a compact token sequence serving as a world prior, and (iii) a shared action expert that attends to this world prior to generate action chunks, with dataset-specific modules aligning state and action interfaces across datasets.

\paperfig{fig:video_generation}{\linewidth}{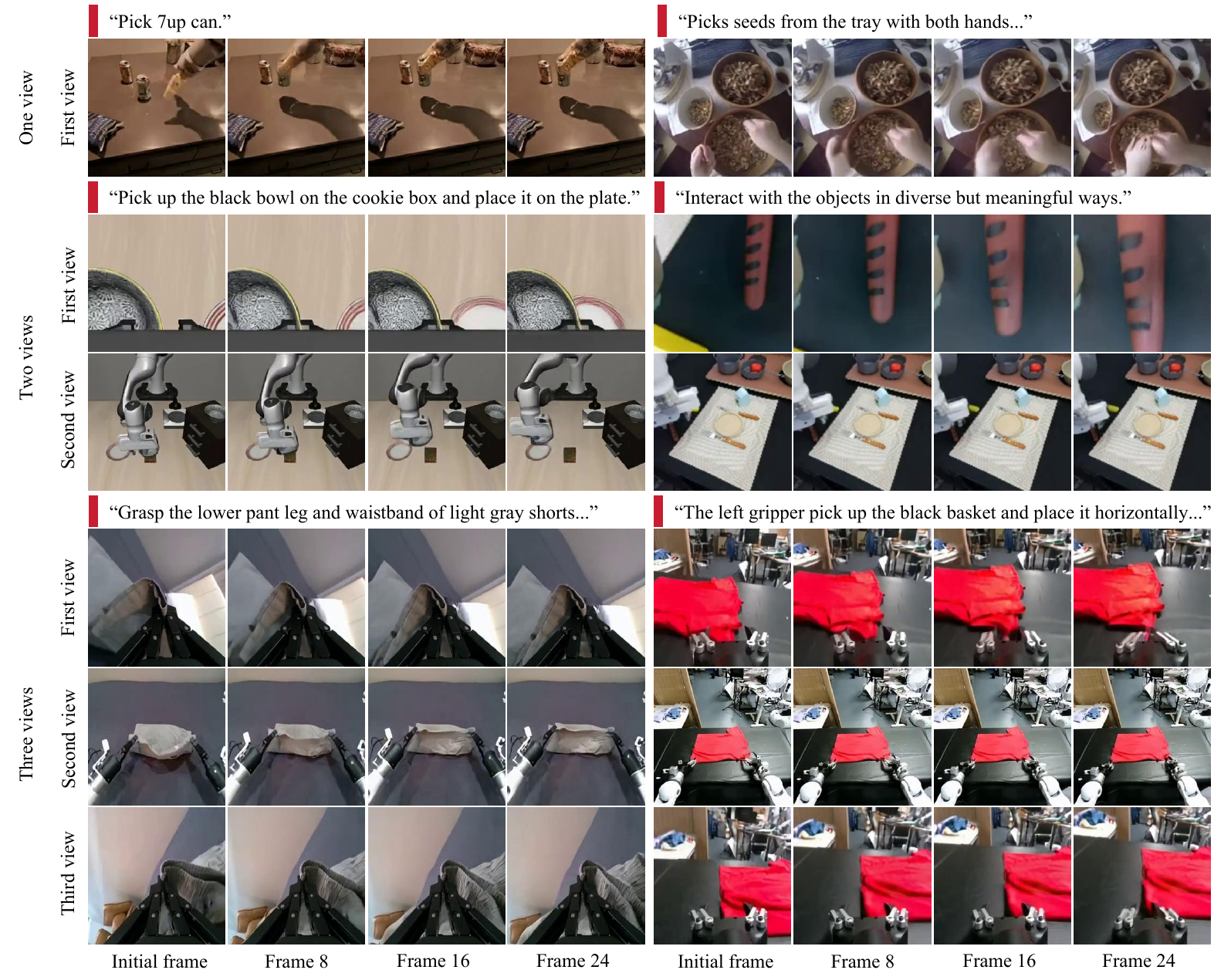}{\textbf{Future-video predictions.} We show predictions at frames 8, 16, and 24. These examples demonstrate that the model supports future-video generation from various views.}

\paperfig{fig:realworld_video_generation}{\linewidth}{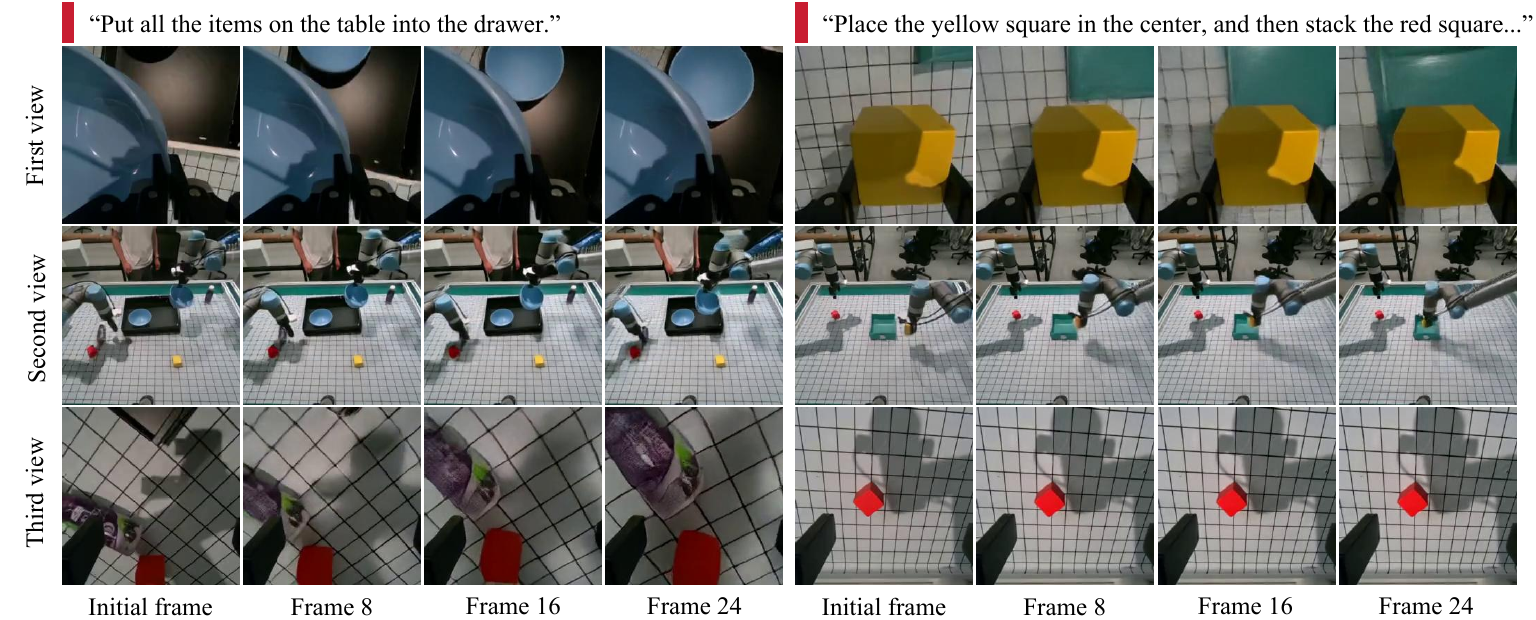}{\textbf{Future-video predictions on real-world teleoperation.} For each of two tasks, we show the observed frame and predictions at frames 8, 16, and 24, illustrating how the model anticipates plausible task progression.}

\noindent\textbf{Video diffusion backbone.}
We start from a pretrained LTX-Video diffusion backbone~\cite{hacohen2024ltx} and adapt it to variable-view robot observations.
Given an instruction $l$ and a \emph{single} variable-view history frame at time $t$, the backbone is trained to predict a $256\times256$ future clip over a video-adaptation horizon of $K{=}24$, distinct from the executable $H{=}16$ action chunk used for closed-loop control.
Its video VAE downsamples by $16\times$ spatially and $8\times$ temporally, reducing token count for efficient training and inference.
Robot setups vary in camera count and viewpoint.
Genie-Envisioner~\cite{liao2025genie} extends LTX-Video to three views by replacing pretrained self-attention with cross-view attention, concatenating tokens across views before attention and splitting them afterward.
However, this modification changes the backbone function at initialization and can shift it away from the pretrained distribution.
Moreover, Genie-Envisioner is trained on the three-view AgiBotWorld dataset~\cite{bu2025agibot}, constraining downstream policies to a three-view observation format.
To preserve the pretrained computation at initialization, we retain per-view self-attention and add cross-view attention as a zero-initialized residual branch (illustrated in the appendix).
We also add a view embedding to encode view identity without assuming a fixed semantic ordering, and we randomize view order during training to improve robustness to heterogeneous camera setups.
Each training sample uses three view slots.
For samples with one or two views, we fill the remaining slots by replicating an existing view and apply independent diffusion noise to each replica, avoiding padding.
A cross-view attention mask blocks interactions between original and replicated slots in both directions.
For different video frame rates, we provide per-sample fps to scale temporal RoPE interpolation~\cite{su2024roformer}, aligning temporal positions with sampling rates.

\noindent\textbf{Q-Former for a compact video world prior.}
Let $\{h_{\ell}\}_{\ell=1}^{L}$ denote the spatiotemporal hidden states from the video backbone's $L$ denoising transformer blocks.
Passing tokens from all video layers directly to the action expert~\cite{liao2025genie,pi05} increases cross-attention memory and leaves the expert to fuse features across layers, which can complicate optimization.
With layerwise conditioning, the expert sees the deepest video features only in its later cross-attention layers.
We distill $\{h_{\ell}\}$ into a fixed-length sequence.
The Q-Former is layer-aligned with the video backbone and refines learnable query tokens through $L$ cross-attention blocks.
At layer $\ell$, the current tokens $z_{\ell-1}$ cross-attend to the corresponding backbone hidden states $h_{\ell}$ and update to $z_{\ell}$.
The final sequence $z\triangleq z_{L}$ is the compact world prior, with a length independent of view and video token counts.

\noindent\textbf{Action expert with dataset-specific alignment.}
VLA datasets have heterogeneous state and action interfaces, including action parameterizations, units, and coordinate frames.
Rather than forcing a single shared action space, inspired by X-VLA~\cite{xvla}, we decouple dataset-specific interfaces from a shared action-generation backbone via lightweight dataset-specific modules.
Each dataset has a state projector, an action input projection for noised action chunks, and an output head that decodes shared representations into executable actions in its native parameterization.
These dataset-specific modules only affect the input and output layers, allowing the shared backbone to scale across heterogeneous state and action interfaces without imposing a single unified action space.
The shared action expert is a DiT-style transformer~\cite{peebles2023scalable} that operates on the concatenated state and action tokens.
It interleaves self-attention over the action-trajectory tokens with cross-attention to the compact world prior $z$ exported by the Q-Former.

\subsection{Training}
\label{subsec:method-training}

We train \method{} in three stages: (1) world prior adaptation, (2) action expert pretraining, and (3) joint post-training on the target-platform dataset.

\noindent\textbf{Stage 1: World prior adaptation.}
We adapt the video diffusion backbone on robot and human videos by optimizing a future-prediction objective in the VAE latent space.
Following standard flow matching, we sample a noise level $\sigma$ from the scheduler and draw $\boldsymbol{\epsilon}\sim\mathcal{N}(0,\mathbf{I})$.
For a clean future latent chunk $\mathbf{y}$, we construct a noised latent $\mathbf{y}_{\sigma}=(1-\sigma)\mathbf{y}+\sigma\boldsymbol{\epsilon}$ and regress the target vector field $(\boldsymbol{\epsilon}-\mathbf{y})$:
\begin{equation}
\mathcal{L}_\mathrm{vid}=\mathbb{E}\Big[\big\|v_{\phi}(\mathbf{y}_{\sigma},\sigma,\mathcal{I}_t,l)-(\boldsymbol{\epsilon}-\mathbf{y})\big\|_2^2\Big].
\end{equation}
To support classifier-free guidance, we apply caption dropout by randomly replacing the instruction with an empty prompt embedding.
We use the variable-view training protocol described in Sec.~\ref{subsec:method-arch} with view-order randomization, view replication for missing view slots, cross-view attention masking, and per-sample fps input for temporal RoPE scaling.
We train on large-scale open-source robot and human video datasets (see the appendix).
We train this stage for 7 days on 64 H100 GPUs.

\noindent\textbf{Stage 2: Action expert pretraining.}
In Stage~2, we freeze the video diffusion backbone and train the Q-Former, the shared action expert, and dataset-specific modules on heterogeneous VLA datasets with action supervision, so this training does not perturb the world prior from Stage~1.
To mitigate dominance from large datasets, we sample training batches using a temperature-scaled mixing scheme to increase the sampling probability of smaller datasets.
Following standard flow matching, we sample an action noise level $\sigma_a$ and draw $\boldsymbol{\epsilon}_a\sim\mathcal{N}(0,\mathbf{I})$.
For a ground-truth chunk $\mathbf{a}\in\mathbb{R}^{H\times d_a}$ with $H{=}16$, we form $\mathbf{a}'=\mathbf{a}+\sigma_a(\boldsymbol{\epsilon}_a-\mathbf{a})$ and regress the target field $(\boldsymbol{\epsilon}_a-\mathbf{a})$:
\begin{equation}
\mathcal{L}_\mathrm{act}=\mathbb{E}\Big[\big\|v_{\theta}(\mathbf{a}',\sigma_a,z,s_t)-(\boldsymbol{\epsilon}_a-\mathbf{a})\big\|_2^2\Big],
\end{equation}
where $d_a$ is the action dimension and $s_t$ is the proprioceptive state.
We train for 3 days on 64 H100 GPUs with large-scale open-source VLA datasets (see the appendix).

\noindent\textbf{Stage 3: Joint post-training.}
Finally, we post-train on self-collected real-world teleoperation trajectories to adapt \method{} to the target platform.
We update all model parameters and optimize a joint objective that combines the video and action losses to adapt both the video world prior and action generation to the target domain:
\begin{equation}
\mathcal{L}_\mathrm{post}=\mathcal{L}_\mathrm{vid}+\mathcal{L}_\mathrm{act}.
\end{equation}

\begin{figure*}[t]
\centering
\begin{minipage}[b]{.322\linewidth}
  \centering
  \includegraphics[width=\linewidth]{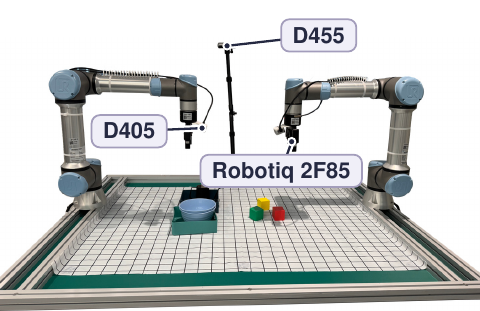}\par
  \footnotesize\textbf{(a)} Side-mounted arms
\end{minipage}\hfill
\begin{minipage}[b]{.322\linewidth}
  \centering
  \includegraphics[width=\linewidth]{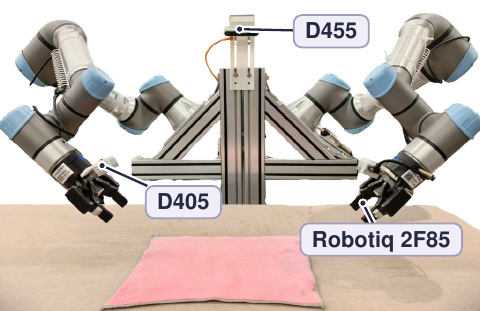}\par
  \footnotesize\textbf{(b)} Two UR5e arms on one frame
\end{minipage}\hfill
\begin{minipage}[b]{.322\linewidth}
  \centering
  \includegraphics[width=\linewidth]{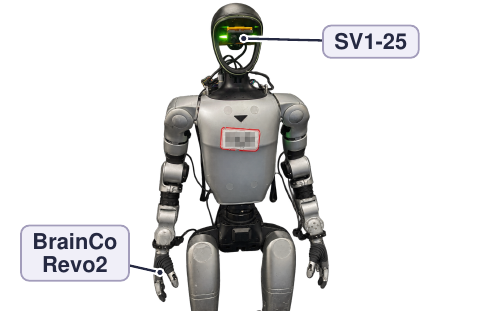}\par
  \footnotesize\textbf{(c)} Unitree G1
\end{minipage}
\caption{\textbf{Real-world robot platforms.} The UR5e setups (a, b) use Robotiq 2F85 grippers, two RealSense D405 wrist cameras, and a D455 third-person camera. The Unitree G1 (c) uses an SV1-25 head camera and BrainCo Revo2 hands.}
\label{fig:real_world_setup}
\end{figure*}

\subsection{Real-time inference}
\label{subsec:method-inference}

At each update, we VAE-encode the current views and initialize future slots with noise. One LTX forward pass extracts hidden states with observed tokens at timestep zero and future tokens at $\sigma{=}1$ (timestep 1000). The Q-Former uses these states to condition a four-step flow-matching action expert.
With asynchronous chunking~\cite{black2025real,black2025training}, the robot executes queued actions while the policy computes their continuation. When the remaining queue length $R$ falls to or below a threshold $P$, the policy uses the latest observations and fixes the $R$ queued actions as the prefix of the new chunk. It drops the overlapping predictions and appends the next $E$ actions to the queue. We choose $P$ so that executing $P$ actions covers the refresh time.
In Stages~2 and~3, we train the expert to continue a fixed prefix by sampling its length uniformly from 0 to 8 and adding noise only to the suffix (see the appendix).

\begin{table}[tb]
\setlength{\abovecaptionskip}{0pt}
\centering
\small
\renewcommand{\arraystretch}{1.05}
\caption{LIBERO success rates (\%) on four suites. Pink marks the best scores (bold) and lavender the second-best (underlined).}
\label{tab:main_results}\label{tab:supp_main_results}
\setlength{\tabcolsep}{2pt}
\begin{tabular*}{\linewidth}{@{\extracolsep{\fill}}>{\raggedright\arraybackslash}p{82pt}cccc|c@{}}
\toprule
Method & Spatial$\uparrow$ & Object$\uparrow$ & Goal$\uparrow$ & Long$\uparrow$ & Average \\
\midrule
Diffusion Policy~\cite{Diffusionpolicy}
& 78.5 & 87.5 & 73.5 & 64.8 & 76.1 \\
OpenVLA~\cite{kim2024openvla}
& 84.7 & 88.4 & 79.2 & 53.7 & 76.5 \\
SpatialVLA~\cite{qu2025spatialvla}
& 88.2 & 89.9 & 78.6 & 55.5 & 78.1 \\
CoT-VLA~\cite{zhao2025cotvla}
& 87.5 & 91.6 & 87.6 & 69.0 & 83.9 \\
$\pi_0$-Fast~\cite{pertsch2025fast}
& 96.4 & 96.8 & 88.6 & 60.2 & 85.5 \\
GR00T-N1~\cite{bjorck2025gr00t}
& 94.4 & 97.6 & 93.0 & 90.6 & 93.9 \\
$\pi_0$~\cite{black2024pi0}
& 98.0 & 96.8 & 94.4 & 88.4 & 94.4 \\
F1~\cite{lv2025f1}
& 98.2 & 97.8 & 95.4 & 91.3 & 95.7 \\
InternVLA-M1~\cite{internvlam1}
& 98.0 & \second{99.0} & 93.8 & 92.6 & 95.9 \\
Genie-Envisioner~\cite{liao2025genie}
& 98.2 & 97.6 & 95.8 & 94.4 & 96.5 \\
$\pi_{0.5}$~\cite{pi05}
& \second{98.8} & 98.2 & \second{98.0} & 92.4 & 96.9 \\
GR00T-N1.6~\cite{gear2025gr00tn16}
& 97.7 & 98.5 & 97.5 & 94.4 & 97.0 \\
OpenVLA-OFT~\cite{OFT}
& 97.6 & 98.4 & 97.9 & 94.5 & 97.1 \\
X-VLA~\cite{xvla}
& 98.2 & 98.6 & 97.8 & \first{97.6} & \second{98.1} \\
\midrule
\method{} & \first{99.6} & \first{99.4} & \first{99.0} & \second{95.8} & \first{98.5} \\
\bottomrule
\end{tabular*}
\end{table}

\section{Experiments}
\label{sec:exp}

\subsection{Video generation}
\label{subsec:exp-video}
We qualitatively evaluate the adapted video diffusion backbone by visualizing its future predictions given a single variable-view history frame and a language instruction.
Figure~\ref{fig:video_generation} shows predictions from the video diffusion backbone after Stage~1 world prior adaptation, while Figure~\ref{fig:realworld_video_generation} shows predictions after further joint post-training with action on $\sim$30 minutes of our real-world teleoperation data.
For each example, we visualize the history frame and the $8$th, $16$th, and $24$th frames from the predicted $K{=}24$ future clip.
Figure~\ref{fig:video_generation} includes examples with one, two, and three input views, highlighting the role of our variable-view design, including view embedding, zero-initialized cross-view attention, and cross-view attention masks.
Figure~\ref{fig:realworld_video_generation} demonstrates that the backbone can be quickly adapted to a new real-world scene with limited data.

\subsection{Simulation results}
\label{subsec:exp-sim}

\paperfig{fig:realworld_vis}{\linewidth}{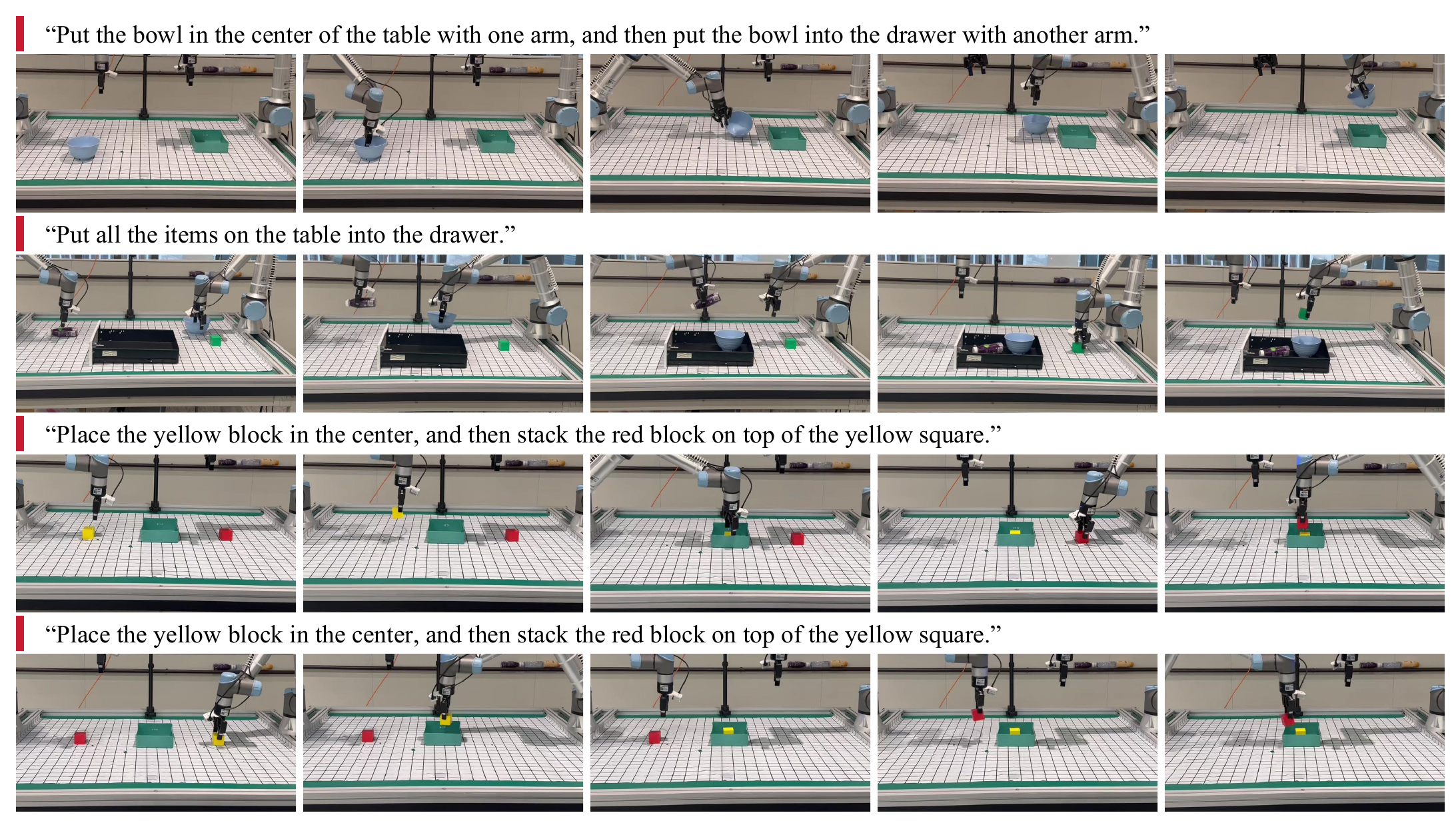}{\textbf{Real-world visualization on the dual-arm UR5e platform.} These rollouts illustrate \method{}'s effective execution across diverse real-world manipulation tasks and its ability to generalize across different initial object layouts.}

\begin{table*}[t]
\setlength{\abovecaptionskip}{0pt}
\centering
\small
\renewcommand{\arraystretch}{1.05}
\caption{Zero-shot success rates (\%) on LIBERO-Plus~\cite{fei2025liberoplus}. Cam., Lang., and Bg. denote camera, language, and background changes, respectively. Best results are bold with pink backgrounds, and second-best results are underlined with lavender backgrounds.}
\label{tab:libero_plus}\label{tab:supp_libero_plus}
\setlength{\tabcolsep}{2pt}
\begin{tabularx}{\linewidth}{@{}>{\raggedright\arraybackslash}p{.24\linewidth}*{8}{>{\centering\arraybackslash}X}}
\toprule
Method & Cam. & Robot & Lang. & Light & Bg. & Noise & Layout & Total \\
\midrule
OpenVLA~\cite{kim2024openvla}
& 0.8 & 3.5 & 23.0 & 8.1 & 34.8 & 15.2 & 28.5 & 15.6 \\
$\pi_0$~\cite{black2024pi0}
& 13.8 & 6.0 & 58.8 & 85.0 & 81.4 & \first{79.0} & 68.9 & 53.6 \\
OpenVLA-OFT~\cite{OFT}
& \second{56.4} & 31.9 & 79.5 & 88.7 & 93.3 & 75.8 & \second{74.2} & 69.6 \\
OpenVLA-OFT$_{\mathrm{w}}$~\cite{OFT}
& 10.4 & 38.7 & 70.5 & 76.8 & \second{93.6} & 49.9 & 69.9 & 55.8 \\
NORA~\cite{hung2025nora}
& 2.2 & 37.0 & 65.1 & 45.7 & 58.6 & 12.8 & 62.1 & 39.0 \\
WorldVLA~\cite{cen2025worldvla}
& 0.1 & 27.9 & 41.6 & 43.7 & 17.1 & 10.9 & 38.0 & 25.0 \\
UniVLA~\cite{bu2025univla}
& 1.8 & 46.2 & 69.6 & 69.0 & 81.0 & 21.2 & 31.9 & 43.9 \\
$\pi_0$-Fast~\cite{pertsch2025fast}
& \first{65.1} & 21.6 & 61.0 & 73.2 & 73.2 & 74.4 & 68.8 & 61.6 \\
RIPT-VLA~\cite{tan2025interactive}
& 55.2 & 31.2 & 77.6 & 88.4 & 91.6 & 73.5 & \second{74.2} & 68.4 \\
OpenVLA-OFT$_{\mathrm{m}}$~\cite{OFT}
& 55.6 & 21.7 & \second{81.0} & \second{92.7} & 91.0 & \second{78.6} & 68.7 & 67.9 \\
X-VLA~\cite{xvla}
& 23.4 & \first{89.7} & 75.7 & 88.2 & \first{96.0} & 62.7 & 71.8 & \second{71.4} \\
\midrule
\method{}
& 52.6 & \second{72.5} & \first{94.3} & \first{95.4} & 85.8 & 59.7 & \first{84.7} & \first{76.5} \\
\bottomrule
\end{tabularx}
\end{table*}

\noindent\textbf{LIBERO.}
We report simulation results on LIBERO~\cite{liu2023libero}, a benchmark for language-instructed tabletop manipulation organized into four suites: Spatial, Object, Goal, and Long.
We post-train a single unified \method{} policy using LIBERO data from all four suites, and evaluate it on each suite by reporting success rate (\%), averaged over 50 episodes for each of the 10 tasks, which amounts to 500 rollouts per suite.
For observations, we use two synchronized RGB views, a third-person view and an eye-in-hand wrist view, along with the language instruction and proprioceptive state.
Table~\ref{tab:main_results} compares \method{} with recent VLA and video-action baselines~\cite{Diffusionpolicy,kim2024openvla,qu2025spatialvla,zhao2025cotvla,pertsch2025fast,bjorck2025gr00t,black2024pi0,lv2025f1,internvlam1,liao2025genie,pi05,OFT,xvla}.
\method{} achieves the highest average success rate of 98.5\%, reaching 99.6\% on Spatial, 99.4\% on Object, and 99.0\% on Goal.
On the Long suite, which stresses long-horizon multi-step execution, \method{} attains 95.8\% success.
Overall, \method{} performs strongly across all four suites, suggesting that transferring the video world prior to action generation improves robustness across diverse task compositions and horizons.

\noindent\textbf{LIBERO-Plus.}
We further evaluate generalization on LIBERO-Plus~\cite{fei2025liberoplus}, which assesses robustness to shifts in camera, robot, language, lighting, background, observation noise, and object layout.
We evaluate the checkpoint post-trained on LIBERO without further adaptation to LIBERO-Plus.
We compare against existing baselines~\cite{kim2024openvla,black2024pi0,OFT,hung2025nora,cen2025worldvla,bu2025univla,pertsch2025fast,tan2025interactive,xvla}.
As shown in Table~\ref{tab:libero_plus}, \method{} reaches a total score of 76.5\%, 5.1 percentage points above X-VLA (71.4\%), the strongest baseline in the table.
It achieves the highest scores under language, lighting, and layout shifts.

\subsection{Real-world results}
\label{subsec:exp-real}
\label{sec:supp-realworld}

\paperfig{fig:realworld_results}{\linewidth}{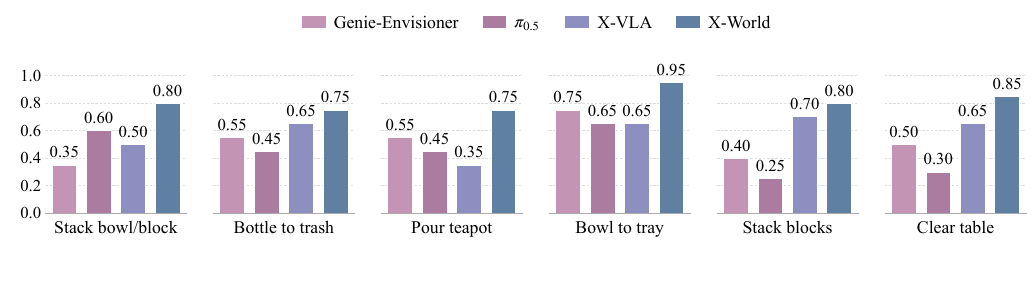}{\textbf{Real-world task success on the side-mounted UR5e setup.} Bars show success fractions over 20 trials per policy and task under a shared evaluation protocol and identical initial placements. \method{} achieves the highest observed success on all six tasks.}

Figure~\ref{fig:real_world_setup} shows three real-robot configurations.
The two dual-arm UR5e setups (a, b) differ in arm mounting. Both use Robotiq 2F85 grippers, RealSense D405 wrist cameras, and a D455 third-person camera.
The Unitree G1 humanoid (c) uses an SV1-25 head camera and BrainCo Revo2 hands for whole-body manipulation.

On the side-mounted UR5e, we evaluate \method{} on six bimanual tasks requiring sequential and simultaneous coordination.
For each task, we collect 30 minutes of VR teleoperation data and post-train the policy on those trajectories.
We record data and run closed-loop control at 30\,Hz, using relative joint position targets for actions and absolute joint positions for proprioception.
We evaluate each policy in 20 trials per task using the same predefined initial placements.

Figure~\ref{fig:realworld_results} reports per-task success rates under a shared evaluation protocol~\cite{liao2025genie,pi05,xvla}.
\method{} achieves the highest success rate on all six tasks, ranging from 75\% to 95\% and averaging 81.7\%.
X-VLA averages 58.3\%, Genie-Envisioner 51.7\%, and $\pi_{0.5}$ 45.0\%.
Across the six tasks, \method{} leads the strongest baseline on each task by 10 to 20 percentage points.

Figure~\ref{fig:realworld_vis} shows representative rollouts.
In our trials, \method{} often recovers from early execution errors by re-localizing in the third-person view. We observe fewer recoveries with X-VLA.
In the training trajectories, the yellow block is always to the left of the red block and is grasped first.
The final rollout shows \method{} following the instructed grasp order after the blocks exchange positions at test time.
The appendix describes the six tasks and presents qualitative examples from the shared-frame UR5e and Unitree G1. Videos of the six side-mounted UR5e tasks are available on the \href{https://logosintelligencelab.github.io/X-World/#demos}{project page}.

\subsection{Ablation studies}
\label{subsec:exp-ablation}

\begin{table}[tb]
\setlength{\abovecaptionskip}{0pt}
\centering
\small
\caption{Ablations of different components on LIBERO.}
\label{tab:ablation}
\setlength{\tabcolsep}{1pt}
\renewcommand{\arraystretch}{1.05}
\begin{tabular*}{\linewidth}{@{\extracolsep{\fill}}lcccc|c@{}}
\toprule
Setting & Spatial$\uparrow$ & Object$\uparrow$ & Goal$\uparrow$ & Long$\uparrow$ & Average \\
\midrule
w/o Stage~1 & 95.4 & 97.6 & 92.4 & 72.4 & 89.5 \\
w/o Stage~2 & 99.2 & 98.8 & 97.2 & 94.2 & 97.4 \\
w/o Stage~1 and Stage~2 & 97.4 & 96.2 & 93.4 & 81.2 & 92.1 \\
Shared projections & 98.6 & 98.4 & 98.2 & 94.4 & 97.4 \\
w/o Q-Former & \first{99.8} & 98.6 & \second{98.6} & 94.6 & 97.9 \\
w/o video loss & \first{99.8} & \first{100.0} & 97.2 & \first{96.2} & \second{98.3} \\
\midrule
Full model & \second{99.6} & \second{99.4} & \first{99.0} & \second{95.8} & \first{98.5} \\
\bottomrule
\end{tabular*}
\end{table}

\noindent\textbf{Training stages.}
We vary Stage~1 (video prior adaptation) and Stage~2 (action expert pretraining) with LIBERO~\cite{liu2023libero} post-training and evaluation fixed.
Variants without Stage~1 start from pretrained LTX-Video~\cite{hacohen2024ltx}, and the variant without either stage goes directly to LIBERO post-training.
Removing Stage~1 from the full pipeline lowers average success from 98.5\% to 89.5\%, with the largest drop on Long (95.8\% to 72.4\%, Table~\ref{tab:ablation}).
Adding Stage~2 raises average success from 97.4\% to 98.5\% with Stage~1 and lowers it from 92.1\% to 89.5\% without Stage~1.

\noindent\textbf{Q-Former.}
Replacing the Q-Former with direct attention to the video backbone's last-layer hidden states lowers average success from 98.5\% to 97.9\%.
Long drops the most (95.8\% to 94.6\%), and Spatial improves slightly.
The comparison jointly varies feature compression and access to intermediate backbone layers.
The Q-Former supplies a fixed-length context to the action expert in 5\,ms.
The direct-attention context scales with the video token count.

\noindent\textbf{Heterogeneous alignment.}
Sharing state and action projections across datasets in Stage~2 lowers average success from 98.5\% to 97.4\%.
All four suites decline, with the largest drop on Long (95.8\% to 94.4\%).

\noindent\textbf{Video loss in post-training.}
Optimizing only the action loss in Stage~3 yields 98.3\% average success versus 98.5\% with joint post-training.
Joint post-training improves Goal (99.0\% vs.\ 97.2\%), while action-only post-training scores higher on Spatial, Object, and Long.

\noindent\textbf{Variable-view design.}
Removing residual cross-view attention or view embeddings lowers average LIBERO success from 98.5\% to 83.1\% or 93.3\%, respectively.
On \textit{Fold Towel}, which requires flattening a cluttered towel before folding it, these variants succeed in 0/20 and 2/20 trials, respectively.
The full policy succeeds in 12/20 trials, and the variant without Stage~2 succeeds in 10/20.
With a single checkpoint, success rises from 40\% with one view to 80\% with three and stays at 80\% when the three views are shuffled (see the appendix).

\begin{table}[tb]
\setlength{\abovecaptionskip}{0pt}
\centering
\small
\renewcommand{\arraystretch}{1.05}
\caption{Inference latency (ms).}
\label{tab:speed_breakdown}\label{tab:runtime}\label{tab:supp_runtime}
\setlength{\tabcolsep}{2pt}
\begin{tabular*}{\linewidth}{@{\extracolsep{\fill}}lr@{\hspace{18pt}}lr@{}}
\toprule
\multicolumn{2}{c}{Model timing} & \multicolumn{2}{c}{Deployment timing} \\
\midrule
T5 encoder (first call) & 13 & Cached model response & 46 \\
VAE & 7 & Network transfer & 32 \\
Video diffusion & 12 & Data processing & 7 \\
Q-Former & 5 & Control-side overhead & 6 \\
Action expert & 8 & End-to-end refresh & 91 \\
\bottomrule
\end{tabular*}%
\end{table}

\noindent\textbf{Runtime and real-time chunking.}
On a single H100, model response takes 59\,ms on the first call and 46\,ms (22\,Hz) with the T5 embedding cached (Table~\ref{tab:speed_breakdown}).
End-to-end refresh takes 91\,ms (11\,Hz).
At 30\,Hz, six committed actions cover 200\,ms, leaving 109\,ms beyond the measured refresh time, and two cover only 67\,ms.
In the prefix-length sweep reported in the appendix, success rises from 65\% at $P{=}2$ to 80\% at $P{=}6$, the value used for deployment.

\section{Conclusion}
\label{sec:conclusion}

We presented \method{}, a world-to-action framework that adapts pretrained video priors to variable-view, language-instructed manipulation.
A Q-Former and flow-matching expert turn video features into action chunks, and asynchronous chunking overlaps inference with execution.
\method{} performs well on LIBERO and zero-shot LIBERO-Plus and averages 81.7\% success across six bimanual tasks on the side-mounted dual-arm UR5e.
\label{sec:main_end}

\section*{LLM Usage Statement}
We used OpenAI Codex to assist with writing and revising project code and
to edit the manuscript for grammar, wording, and clarity.

\subsection*{Ethics statement}

This work uses public human and robot video datasets and self-collected robot
teleoperation data for research on language-instructed manipulation. Public
data are used according to their licenses. Teleoperation recordings are
handled as potentially identifying data, and any redistribution should
preserve the applicable consent and privacy requirements. Physical robot
experiments require supervised operation, workspace safeguards, and an
emergency stop; success in simulation or a qualitative rollout does not
establish general hardware safety.

\subsection*{Reproducibility statement}

The paper reports the training objectives, updated modules, input resolution,
prediction horizons, feature-extraction path, and evaluation settings.
The Appendix lists data sources, frame-rate interpolation, native action
representations, and the available aggregate results.
Figure scripts use those aggregates and the original qualitative frames.
Full model dimensions, optimization configurations, post-training budgets,
and per-run evaluation logs are not supplied here; these omissions limit
independent reproduction and uncertainty analysis.

{
  \small
  \bibliographystyle{ieeenat_fullname}
  \bibliography{main}
}

\clearpage
\appendix
\setcounter{figure}{0}
\setcounter{table}{0}
\setcounter{equation}{0}
\renewcommand{\thefigure}{S\arabic{figure}}
\renewcommand{\thetable}{S\arabic{table}}
\renewcommand{\theequation}{S\arabic{equation}}
\renewcommand{\theHfigure}{supp.\arabic{figure}}
\renewcommand{\theHtable}{supp.\arabic{table}}
\renewcommand{\theHequation}{supp.\arabic{equation}}

\twocolumn[{%
  {\centering\Large\bfseries Supplementary Material\par}
  \vspace{1em}
  \noindent\begin{minipage}{\linewidth}
\captionsetup{type=figure}
\centering
\includegraphics[width=\linewidth]{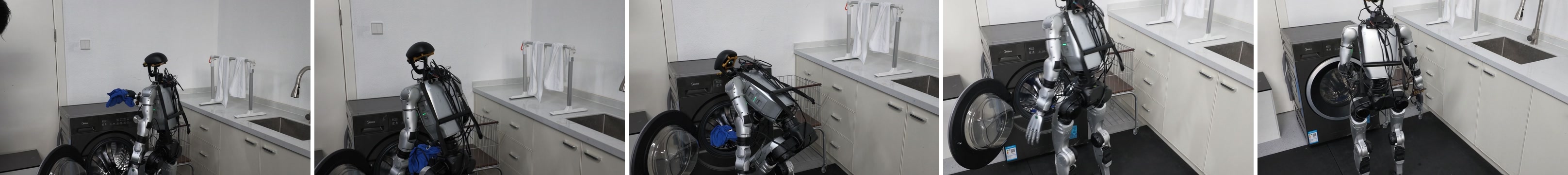}\par\nointerlineskip\vspace{1pt}
\includegraphics[width=\linewidth]{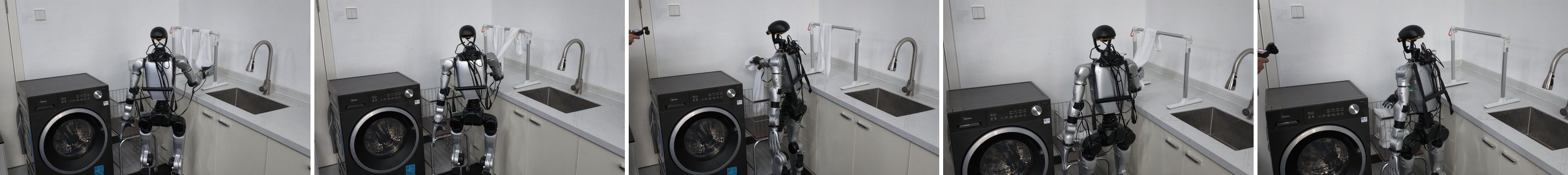}\par\nointerlineskip\vspace{1pt}
\includegraphics[width=\linewidth]{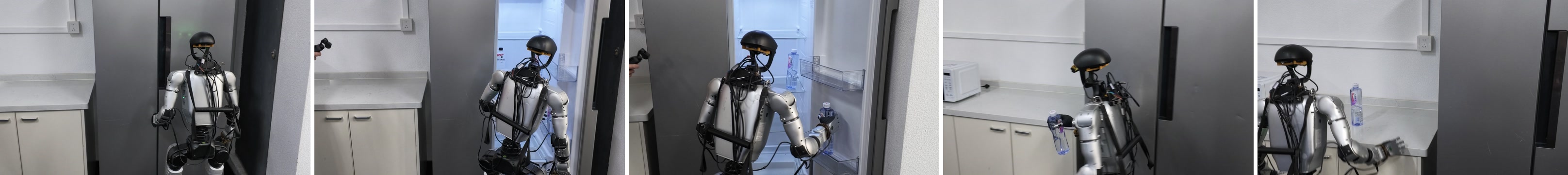}\par\nointerlineskip\vspace{1pt}
\includegraphics[width=\linewidth]{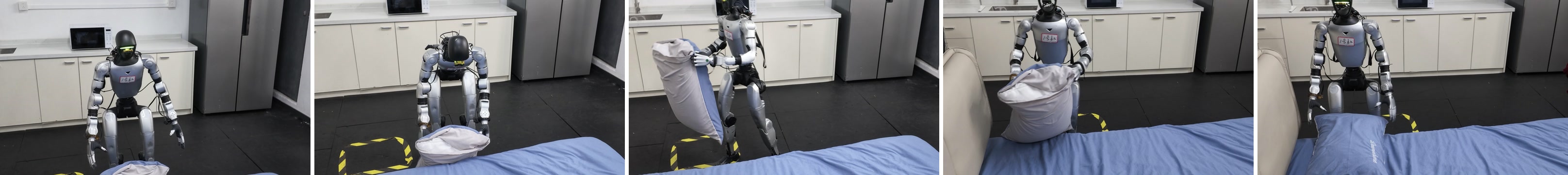}\par
\caption{\textbf{Household task sequences on Unitree G1.}
From top to bottom, the rows show loading a towel into a washer, putting white towels in a basket,
moving a water bottle from the fridge to the counter, and placing a pillow on a bed. Time runs left to right.}
\label{fig:additional_realworld_examples}
\end{minipage}\par

  \vspace{0.7\baselineskip}
}]
\begin{figure*}[t]
\centering
\includegraphics[width=\linewidth]{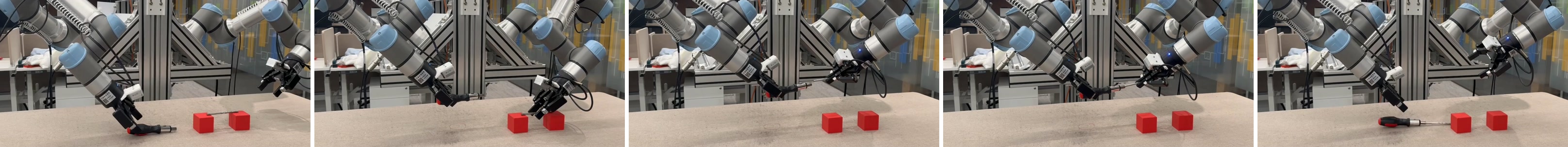}\par\nointerlineskip\vspace{1pt}
\includegraphics[width=\linewidth]{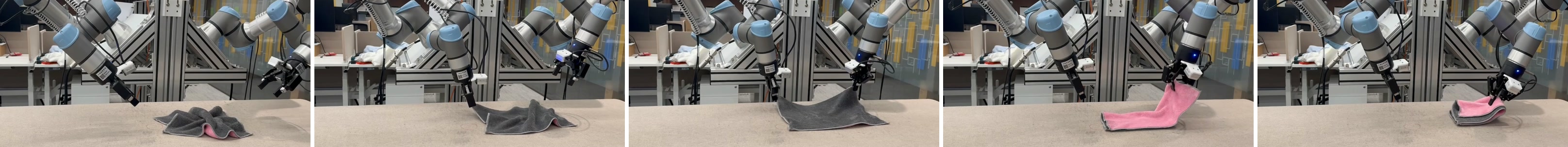}\par\nointerlineskip\vspace{1pt}
\includegraphics[width=\linewidth]{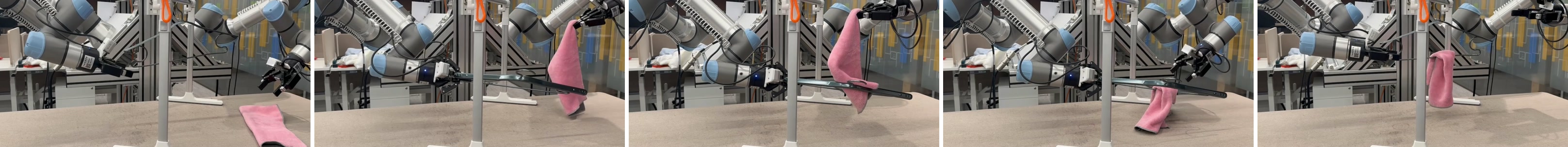}\par\nointerlineskip\vspace{1pt}
\includegraphics[width=\linewidth]{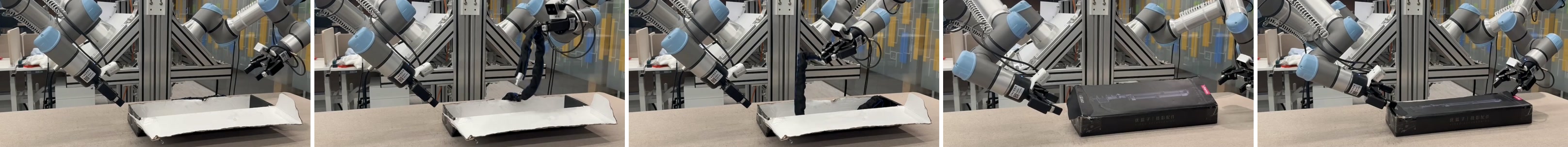}\par
\caption{\textbf{Task sequences on the shared-frame dual-arm UR5e.}
The rows show bit insertion, towel folding, towel hanging, and chain packing from top to bottom.
Time runs from left to right.}
\label{fig:shared-frame-rollouts}
\end{figure*}

\section{Additional Experiments}
\label{sec:supp-additional-ablation}

\subsection{Real-Robot Examples}
\label{sec:supp-robot-examples}

Videos of the six side-mounted dual-arm UR5e tasks are available on the \href{https://logosintelligencelab.github.io/X-World/#demos}{project page}.
Figures~\ref{fig:additional_realworld_examples} and~\ref{fig:shared-frame-rollouts} show task sequences for the Unitree G1 and the shared-frame dual-arm UR5e, respectively.

\begin{figure}[!t]
\centering
\includegraphics[trim=0 23bp 0 14bp,clip,width=\linewidth]{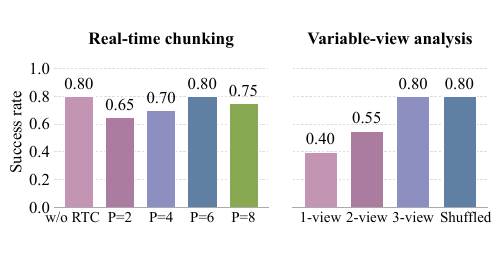}
\caption{\textbf{Action-prefix and view-availability analyses.} The left panel compares inference without real-time chunking (w/o RTC) with prefix lengths~\mbox{$P=2,4,6,8$}. The right panel uses the same checkpoint to compare one, two, and three camera views and a shuffled three-view order. Bars report success fractions.}\label{fig:supp_additional_results}
\end{figure}

\subsection{Asynchronous real-time chunking}

Figure~\ref{fig:supp_additional_results} (left) compares asynchronous chunking at $P=2,4,6,8$ with execution without RTC. Success is 65\%, 70\%, 80\%, and 75\% for those prefixes. Without RTC, success is 80\%, and execution pauses while the policy refreshes. Short prefixes give the policy less time to refresh the queue, while long prefixes reduce update frequency. We use $P{=}6$ in the real-world experiments.

\subsection{Variable-view analysis}

We use the same checkpoint across all view settings. The one-view setting retains the third-person camera, the two-view setting removes one wrist camera, and the shuffled setting permutes the three camera inputs. Figure~\ref{fig:supp_additional_results} (right) shows success rates of 40\%, 55\%, 80\%, and 80\% for one, two, three, and shuffled views, respectively. Success increases with view count and stays at 80\% under the tested permutation.

\section{Data and Training Details}
\label{sec:supp-data}

This section summarizes the data sources used across the three training stages of \method{} and the sampling strategy used for heterogeneous action pretraining.

\subsection{Stage 1: Video pre-training}
\label{sec:supp-data-stage1}

We use a mixture of robot videos and human videos to provide diverse visual interaction data for video pre-training. All sources are converted into a unified variable-view format, with up to three views retained for each sample. Table~\ref{tab:supp_data_stage1} summarizes the resulting data sources.

\subsection{Stage 2: Action pre-training}
\label{sec:supp-data-stage2}

We use heterogeneous VLA trajectories collected from multiple benchmarks and robot platforms for action pre-training. These datasets differ in embodiment, observation setup, and heterogeneous state and action interfaces, providing broad supervision across diverse manipulation settings. Table~\ref{tab:supp_data_stage2} summarizes the main VLA data sources.

\subsection{Temperature-scaled multi-dataset sampling}
\label{sec:supp-data-mixing}

To avoid over-sampling the largest sources while still accounting for dataset scale, we use temperature-scaled sampling when drawing training examples from the Stage~2 mixture. The sampling weight of dataset $d$ is defined as
\begin{equation}
w_d\propto \alpha_d\cdot N_d^{1/T},
\end{equation}
where $N_d$ denotes the total number of dataset steps, i.e., the number of sampleable state and action transitions in dataset $d$, $\alpha_d$ is the base mixture weight, and $T$ is the sampling temperature. We set $\alpha_d=1.0$ and use $T=2.0$ for all datasets, so the effective reweighting scales with the square root of the dataset step count before normalization.

\subsection{Stage 3: Real-world post-training data}
\label{sec:supp-data-stage3}

We use self-collected teleoperation data for post-training on the six real-world tasks in \mainsectionref{sec:supp-realworld}. We collect 30 minutes of demonstrations per task with synchronized three-view observations.

\noindent\textbf{Real-world task descriptions.}
\textit{Stack bowl and block} stacks a side bowl onto a center bowl before placing a block into it.
\textit{Put bottle into trash can} picks up scattered bottles and places them in a trash can.
\textit{Pour water from teapot} pours water into a cup before placing the teapot and cup on a tray.
\textit{Move bowl into a tray} moves a bowl to the center with one arm and transfers it into a tray with the other.
\textit{Stack two blocks} moves two blocks from the table to a tray and stacks them.
\textit{Clear the table} places all objects into a drawer.

\begin{table*}[ht]
\centering
\small
\caption{Datasets for Stage~1 video adaptation. Robot and human videos supply future-frame supervision, while Views gives the number of available camera streams per training example.}
\label{tab:supp_data_stage1}
\setlength{\tabcolsep}{2.2pt}
\renewcommand{\arraystretch}{1.01}
\begin{tabularx}{\linewidth}{@{}>{\raggedright\arraybackslash}p{0.28\linewidth}c>{\raggedright\arraybackslash}X@{}}
\toprule
Dataset & Views & Summary \\
\midrule
RoboTwin 2.0~\cite{chen2025robotwin} & 3 & Domain-randomized dual-arm benchmark designed for robust bimanual manipulation and transfer. \\
CALVIN~\cite{mees2022calvin} & 2 & Long-horizon language benchmark built around static and gripper observations for sequential manipulation. \\
AgiBotWorld~\cite{bu2025agibot} & 3 & Quality-controlled manipulation data collected in realistic settings with broad skill coverage. \\
Behavior-1K~\cite{li2024behavior1k} & 3 & Household simulation benchmark emphasizing realistic environments and long-horizon daily activities. \\
RoboCOIN~\cite{wu2025robocoin} & 3 & Bimanual multi-embodiment data with hierarchical annotations for coordinated manipulation skills. \\
Galaxea~\cite{jiang2025galaxea} & 3 & Open-world robot data from everyday scenes with fine-grained subtask language annotations. \\
Open-X-Embodiment~\cite{o2024open} & 1-3 & Aggregated real-robot data spanning diverse embodiments, camera layouts, and manipulation behaviors. \\
RoboMIND~\cite{wu2024robomind} & 1-3 & Unified multi-embodiment teleoperation data covering diverse tasks, objects, and robot platforms. \\
Ego4D~\cite{grauman2022ego4d} & 1 & Large-scale egocentric daily-life videos capturing diverse activities and environments from a first-person view. \\
Humanoid-Everyday~\cite{zhao2025humanoid} & 1 & Open-world humanoid manipulation data with multimodal sensing for whole-body interaction tasks. \\
\bottomrule
\end{tabularx}
\end{table*}

\begin{table*}[ht]
\centering
\small
\caption{Datasets for Stage~2 action pretraining. Robot trajectories provide action supervision, while Views gives the available camera count per example; dataset-specific projections accommodate native state and action formats.}
\label{tab:supp_data_stage2}
\setlength{\tabcolsep}{2.2pt}
\renewcommand{\arraystretch}{1.01}
\begin{tabularx}{\linewidth}{@{}>{\raggedright\arraybackslash}p{0.28\linewidth}c>{\raggedright\arraybackslash}X@{}}
\toprule
Dataset & Views & Summary \\
\midrule
RoboTwin 2.0~\cite{chen2025robotwin} & 3 & Domain-randomized dual-arm trajectories for robust bimanual manipulation and transfer. \\
CALVIN~\cite{mees2022calvin} & 2 & Long-horizon language-conditioned trajectories with static and gripper observations. \\
AgiBotWorld~\cite{bu2025agibot} & 3 & Large-scale manipulation trajectories collected in realistic settings with diverse skills. \\
Behavior-1K~\cite{li2024behavior1k} & 3 & Household embodied benchmark providing long-horizon trajectories in realistic simulation. \\
RoboCOIN~\cite{wu2025robocoin} & 3 & Bimanual multi-embodiment trajectories with hierarchical annotations for coordinated skills. \\
Galaxea~\cite{jiang2025galaxea} & 3 & Open-world manipulation trajectories with fine-grained subtask language supervision. \\
Open-X-Embodiment~\cite{o2024open} & 1-3 & Aggregated cross-embodiment trajectories covering diverse robots, scenes, and tasks. \\
RoboMIND~\cite{wu2024robomind} & 1-3 & Unified multi-embodiment teleoperation trajectories across diverse tasks and platforms. \\
\bottomrule
\end{tabularx}
\end{table*}

\section{Implementation Details}
\label{sec:supp-impl}

This section provides additional implementation details for variable-view processing, the token sequence interface, heterogeneous state and action alignment, and asynchronous real-time chunking.

\subsection{Variable-view video processing}
\label{sec:supp-impl-variable-view}
\begin{figure*}[t]
\centering
\includegraphics[width=\linewidth]{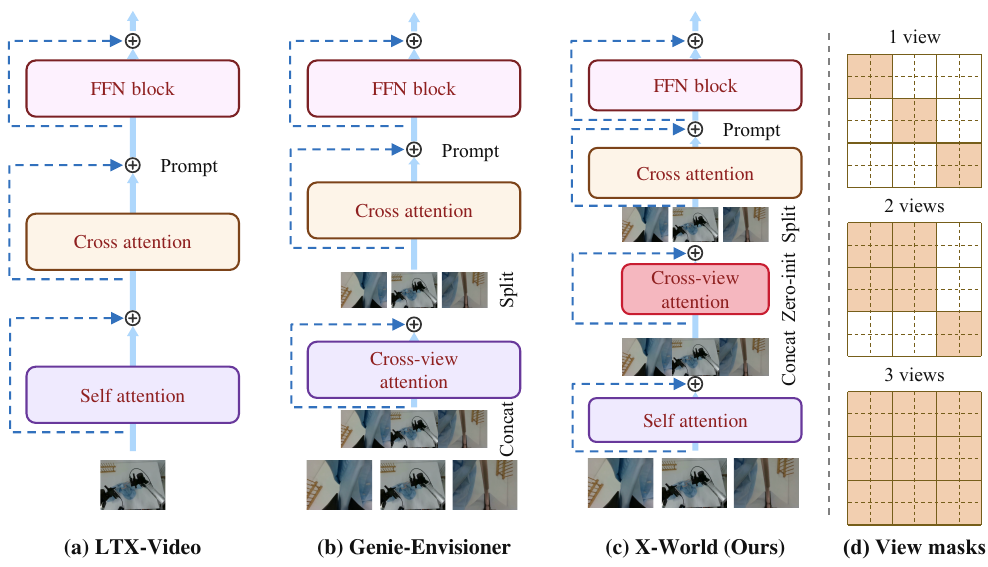}\par
\setlength{\abovecaptionskip}{4pt}
\caption{\textbf{Cross-view adaptation of the video transformer.} (a)~LTX-Video~\cite{hacohen2024ltx} applies self-attention to a single view. (b)~Genie-Envisioner~\cite{liao2025genie} fuses concatenated views through cross-view attention. (c)~\method{} retains pretrained per-view self-attention and learns cross-view fusion through a separate residual branch. Zero initialization preserves the pretrained computation at the start of adaptation while allowing cross-camera interactions to be learned. (d) Training-time masks for one, two, and three available views. Shaded entries allow attention; replicated slots are isolated from available views, supporting joint training across camera counts.}
\label{fig:genie_mask}
\end{figure*}

We add a view embedding to encode view identity without assuming a fixed semantic ordering, and randomize view order during training to improve robustness to heterogeneous camera setups. Each training sample is represented with three view slots. When only one or two views are available, we replicate an existing view to fill the missing slots and apply independent diffusion noise to each replica, which avoids padding and prevents wasting computation on masked placeholders. Cross-view attention masks block interactions between replicated and original view slots, ensuring that replicated views do not affect the original views and vice versa.
Figure~\ref{fig:genie_mask} illustrates the cross-view residual branch and the attention masks for one to three available views.

\subsection{FPS-aware temporal RoPE scaling}
\label{sec:supp-impl-fps}

Our video diffusion backbone builds on LTX-Video~\cite{hacohen2024ltx} and ingests videos collected at different frame rates. To align temporal positional encoding with the physical sampling rate of each sample, we pass the per-sample frame rate $\mathrm{fps}$ to the backbone and use it to construct the RoPE interpolation tuple~\cite{su2024roformer} in both training and inference:
\begin{equation}
\mathrm{rope\_interpolation\_scale}=\Big(\tfrac{r_t}{\mathrm{fps}},\ r_s,\ r_s\Big),
\end{equation}
where $r_t$ and $r_s$ denote the temporal and spatial compression ratios of the LTX-Video VAE. For multi-view inputs, the same sample-level $\mathrm{fps}$ is repeated across views. In the transformer, this tuple scales the temporal, height, and width coordinates before rotary embedding is applied, so clips with different frame rates share the same architecture while their temporal coordinates remain aligned with the actual sampling rate.

\subsection{Q-Former and token sequence interface}
\label{sec:supp-impl-qformer}

The video diffusion backbone produces spatiotemporal features throughout its denoising transformer blocks. We denote the exported hidden states by $\{h_{\ell}\}_{\ell=1}^{L}$ and distill them into a fixed-length token sequence using a layer-aligned Q-Former. At layer $\ell$, the current query tokens cross-attend to the corresponding backbone hidden states $h_{\ell}$ before being passed to the next layer. The final token sequence forms a compact world prior whose size is independent of the number of input views and video tokens, and the action expert attends to this fixed interface when generating action chunks.

\subsection{Action representation and dataset-specific modules}
\label{sec:supp-impl-action}

VLA datasets have heterogeneous state and action interfaces, including different action parameterizations, units, and coordinate frames. Rather than forcing a single shared action space, we use lightweight dataset-specific modules around a shared action-generation backbone, following the design in X-VLA~\cite{xvla}. Each dataset has a state projector, an action input projection for noised action chunks, and an output head that decodes shared representations into executable actions in its native parameterization. For the real-world platform, actions are represented as relative joint position targets, while absolute joint positions are provided as proprioceptive state.

\subsection{Asynchronous real-time chunking}
\label{sec:supp-impl-chunking}

We maintain a FIFO action queue $\mathcal{Q}$ and execute and remove its front action at each control step. Let $R\triangleq |\mathcal{Q}|$ be the remaining queue length, $P$ the prefix threshold, and $E$ the number of newly predicted suffix actions appended after each refresh. When $R\leq P$, the policy uses the current observations and reuses the remaining $R$ actions as a committed prefix to generate a refreshed action chunk. It discards the overlapping $R$ predicted actions and appends the next $E$ suffix actions to $\mathcal{Q}$. The robot continues executing committed actions while the policy updates future actions in the background, decoupling inference updates from action execution and avoiding waiting gaps.

In the reference deployment implementation, $P$ corresponds to prefix\_steps, $E$ corresponds to execution\_steps, and delay is computed as $\min(\text{remain\_steps},\text{prefix\_steps})$. Thus, delay gives the number of remaining queued actions reused as the action prefix and dropped from the overlapping part of the returned prediction.

In Stages~2 and~3, we sample the prefix length $\delta$ uniformly from $\{0,\ldots,8\}$ and set the per-step action noise level to
\begin{equation}
\sigma_{a,i} = \begin{cases}
0, & i\leq\delta,\\
\sigma_a, & i>\delta,
\end{cases}
\end{equation}
where $\sigma_a$ is the sampled action noise level. We form $\mathbf{a}'_i=\mathbf{a}_i+\sigma_{a,i}(\boldsymbol{\epsilon}_{a,i}-\mathbf{a}_i)$ with $\boldsymbol{\epsilon}_{a,i}\sim\mathcal{N}(0,\mathbf{I})$.

\section{Limitations}

Transfer to unseen robot embodiments remains an open question.
To adapt to a new robot, we would need to pretrain the action expert again with data collected on that robot.
We plan to test this approach on robots outside our current training set.

\end{document}